\documentclass[journal]{IEEEtran}  

\IEEEoverridecommandlockouts                              

\usepackage{graphicx} 
\usepackage{cite}
\usepackage{amsmath,amssymb,amsfonts}
\usepackage{algorithmic}
\usepackage{graphicx}
\usepackage{textcomp}
\usepackage{mathrsfs}
\usepackage{epsfig}
\usepackage{mathptmx}
\usepackage{times}
\usepackage{mathtools}
\usepackage{booktabs}
\usepackage{placeins}

\usepackage{color,soul}
\usepackage{algorithm}
\usepackage{algorithmic}

\floatname{algorithm}{Implementation of the Simulation Studies}

\title{\LARGE \bf Robotic Knee Tracking Control to Mimic the Intact Human Knee Profile Based on Actor-critic Reinforcement Learning}

\author{Ruofan Wu, Zhikai Yao, Jennie Si, \textit{Fellow, IEEE} and He (Helen) Huang, \textit{Senior Member, IEEE}
\thanks{This work was partly supported by National Science Foundation: \#1563921 and \#1808752, \#1563454 and \#1808898.}
\thanks{R. Wu$^*$, Z. Yao$^*$ and J. Si are with the School of Electrical, Computer
	and Energy Engineering, Arizona State University, Tempe, AZ 85287 USA, $^*$equal contribution, e-mail: ruofanwu@asu.edu; zacyao.cn@gmail.com; si@asu.edu (correspondence).} 
\thanks{H. Huang is with the Department of Biomedical Engineering, North Carolina State University, Raleigh, NC 27695 USA, and also with the University of North Carolina at Chapel Hill, Chapel Hill, NC 27599 USA (e-mail: hhuang11@ncsu.edu).}
}

\begin{document}

\maketitle
\thispagestyle{empty}
\pagestyle{empty}

\begin{abstract}
We address a state-of-the-art reinforcement learning (RL) control approach to automatically configure robotic prosthesis impedance parameters to enable end-to-end, continuous locomotion intended for transfemoral amputee subjects. Specifically, our actor-critic based RL provides tracking control of a robotic knee prosthesis to mimic the intact knee profile. This is a significant advance from our previous RL based automatic tuning of prosthesis control parameters which have centered on regulation control with a designer prescribed robotic knee profile as the target. In addition to presenting the complete tracking control algorithm based on direct heuristic dynamic programming (dHDP), we provide an analytical framework for the tracking controller with constrained inputs. We show that our proposed tracking control possesses several important properties, such as weight convergence of the learning networks, Bellman (sub)optimality of the cost-to-go value function and control input, and practical stability of the human-robot system under input constraint. We further provide a systematic simulation of the proposed tracking control using a realistic human-robot system simulator, the OpenSim, to emulate how the dHDP enables level ground walking, walking on different terrains and at different paces. These results show that our proposed dHDP based tracking control is not only theoretically suitable, but also practically useful.

\end{abstract}

\begin{figure*}[!htp]
	\centering
	\includegraphics[width=380pt]{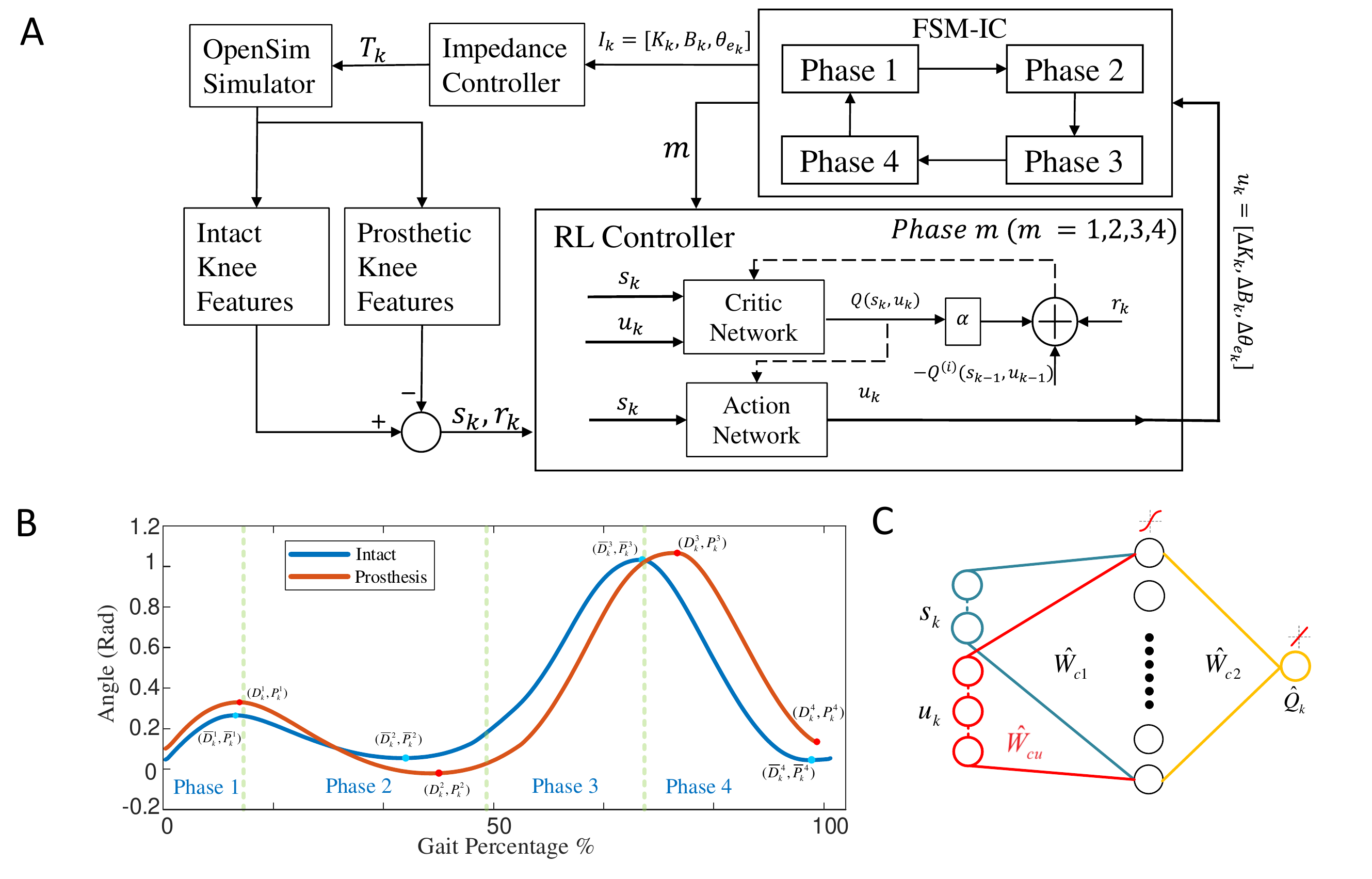}\\
	\caption{A) Block diagram of the automatic robotic knee control parameter tuning scheme by dHDP. Two loops operate at two different time scales: 1) the impedance controller provides outputs at 100 Hz to regulate the joint torque; 2) the dHDP is updated every step (denoted by $k$) to adjust the impedance parameters (4). B) Complete gait profiles of the intact and robotic knee, where $Y^m_k=(D^m_k,P^m_k)$ is the intact knee as target position and $Z^m_k=(D^m_k,P^m_k)$ the robotic knee position, m=1, 2, 3, 4, $D_k$ the gait percentage and $P_k$ the knee angle. C) The critic network.}
	\label{fig:experiment}
\end{figure*} 


\section{INTRODUCTION}

Powered lower limb prosthesis provides great promise for amputees to regain mobility in daily life. Its potential has been demonstrated for transfemoral amputees’ walking ability \cite{martinez2009agonist, johansson2005clinical}. Such robotic devices rely on an impedance control framework which is designed based on human  biomechanics to mimic the central nervous system controlled human joint movements to provide a natural substitute to the lost limb functions. These devices requires customization of the impedance parameters for each individual user. Currently, configuration of the powered devices is performed in clinics by technicians who manually tune a subset of impedance parameters over a number of visits of the patient. This procedure is time and labor intensive for both amputees and clinicians. Therefore, an automatic approach to tuning the powered prosthesis parameters is needed.

Automatically configuring the impedance parameter settings has been attempted over the past several years. An untested idea aims at estimating the joint impedance based on biomechanical measurements and a model of the unimpaired leg \cite{ rouse2014estimation ,pfeifer2012model }. This idea may not be practically useful as the biomechanics and the joint activities of amputees are fundamentally different from those of the able-bodied population. Another approach is to constrain the knee kinematics via the relationship of the joint control and intrinsic measurements, which in turn requires careful modeling and thus may not be feasible \cite{ gregg2013experimental, eilenberg2010control}. Such an approach relies on significant domain knowledge and is tuning time. A cyber expert system was proposed\cite{huang2016cyber} to emulate the prosthetists’ tuning decisions of human experts into configuring the control parameters. This approach heavily relies on the expert’s experience and is not expected to scale well to more joints and different users and tasks. 

As those methods all have their fundamental limitations in principled ways, new approaches to configuring the prosthesis control parameters are needed. RL based adaptive optimal control approaches is a promising alternative as they have demonstrated their capability of learning from data measurements in an online or offline manner in several realistic application problems including large-scale control problems\cite{lu2008direct, guo2015approximate, guo2015online, enns2000helicopter, enns2002apache, enns2003helicopter}. The core of the RL methods is the idea of providing approximate solutions to the Bellman equation of optimal control problems. We have successfully developed several RL algorithms to configure impedance parameter settings, including actor-critic RL \cite{wen2018robotic,Wen2017,wen2019online,wen2020} and policy iteration based RL approaches\cite{ gao2018robotic, Li2019,li2020towards,gao2020reinforcement}, and systematically tested them in both extensive simulations and in experiments using able-bodied and transfemoral amputee subjects. 

All of our RL control approaches to date require a target knee motion profile which can only be subjectively determined. Nonetheless, those results are important as they provided the necessary understanding of the impedance control parameter configuration problem and if RL control is capable of solving this problem. Even though detailed understanding of human locomotion at neurological and biomechanical levels has long been established, an individual human subject's locomotion dynamics are still not feasible to model accurately by mathematical descriptions as individuals differ physically, biologically and neurologically. Additionally, different locomotion tasks, such as changing pace \cite{pietraszewski2012three}, sloped walking \cite{lay2006effects} and walking on uneven terrain \cite{voloshina2013biomechanics} all have significant influence on human gait behavior. As such, accurately prescribing each and every locomotion behavior for control purposes is not feasible. 
 
Tracking the intact knee joint motion by a prosthetic knee is an intuitive idea as the intact knee kinematics is the most natural and realistic target: it contains actual biological joints’ inter-relational information, which makes it a good candidate to replace a subjectively defined knee profile. Studies have shown that bilateral coordination between two legs are needed in the regulation of bipedal walking to maintain stability, and that such interlimb cooperation can be accomplished at a spinal level. Since the spinal level locomotor network are symmetrically organized\cite{d2014modulation}, sensory and muscle activity of both sides are involved in rhythmic walking. Amputees usually display asymmetrical walking by relying heavily on their intact limb because of the loss of sensory feedback. 
Tracking the intact knee actually has been explored years ago. Grimes et al. developed a mirror control scheme for the stance phase (not a complete gait cycle). It tracked the sound limb’s knee trajectory in the stance phase by multiplying a gain factor to avoid over flexion while a fixed trajectory was applied in swing phase. Melek et al. copied the full gait trajectory by Kalman filter with a biomimetic designed prosthesis but no human experiment or systematic simulations were reported\cite{bernal2018design}. Joshi et al. developed a control strategy by controlling the swing time to mirror the stride duration of the intact knee while the prothesis was locked during stance phase\cite{Joshi2010}. Sahoo et al. aimed at mirroring the step length by controlling the push-off force\cite{sahoo2018novel}. The above approaches either focused on part of a gait cycle or the outcome measurement such as step length and stance time. None of them has shown feasibility of tracking a completed gait cycle.

Virtual constraints were proposed to generate coordinated joint motions as target joint motion profiles for the robotic knee to track\cite{kumar2019extremum}. Biomimetic virtual constraints described the joints’ geometric relationships and were encoded by a hybrid zero dynamics \cite{westervelt2003hybrid}. However, there are a few limitations on this approach. Virtual constraints require a simplified human model to establish the geometric relationship among joints. Such a model is difficult to establish for a human-prosthesis system. In a recent work of prosthesis control design based on the virtual constraints\cite{kumar2019extremum}, only a proportional gain was derived and applied. The overall human-prosthesis performance during locomotion is yet to be demonstrated.

In this paper, we propose a RL tracking control scheme for the robotic knee to mimic the intact knee in different locomotion tasks.  In a previous experiment \cite{wu2021reinforcement}, we successfully tested this pilot idea of RL tracking control to automatically configure impedance parameter settings. In this study, we formally  formulate the tracking control problem, develop a complete tracking control algorithm based on dHDP, and provide a analytical framework to validate the real time control performance guarantee by using this proposed scheme. The contributions of this work include the following. 

1) We provide a new, RL based tracking control solution of a robotic knee prosthesis to mimic the intact knee profile. This is the first systematic demonstration of an end-to-end, continuous walking enabled by automatic tracking control of a wearable lower limb robotic device. All our previous results to date \cite{wen2018robotic,Wen2017,wen2019online,wen2020,gao2018robotic, Li2019,li2020towards,gao2020reinforcement} are based on regulation control with the desired prosthetic knee profile prescribed. 

2)	We provide an analytical framework for a constrained input tracking control of the prosthetic knee. Based on a successful actor-critic learning control algorithm, the dHDP, we show that our proposed tracking control possesses several important properties, such as weight convergence of the learning networks, Bellman (sub)optimality of the cost-to-go value function and control input, and practical stability of the human-robot system under input constraint.

3)  We provide a systematic evaluation of the proposed tracking control using a realistic human-robot system simulator, the OpenSim, to demonstrate that the dHDP enabled robotic knee can successfully track the intact knee profile for level ground walking, slope walking under various slope angles and walking under different paces. These results show that our proposed dHDP based tracking control is not only theoretically suitable, but also practically useful.

The remaining of this paper is organized as follows. Section \uppercase\expandafter{\romannumeral2} describes the human-prosthesis system and develops dHDP to solve the tracking problem. Section \uppercase\expandafter{\romannumeral3} gives Lyapunov stability analysis of system. Section \uppercase\expandafter{\romannumeral4} presents the implementation using Opensim simulations. Section \uppercase\expandafter{\romannumeral5} presents extensive simulations.  Discussions and conclusion are presented in Section \uppercase\expandafter{\romannumeral6}.

\section{Method}

Our proposed RL tracking control is built upon the finite state machine (FSM) impedance controller (IC) framework. It is to mimic the torque-generating capability of biological joints to enable natural movement. The FSM-IC provides intrinsic control in the form of adjustable control torque influenced by impedance parameters. The settings of the impedance parameters as control inputs have to be adjusted or adapted to meet individuals' needs including their different physical condition. Our proposed RL tracking control is to automatically provide such needed impedance parameter settings.

\subsection{Finite State Machine (FSM) Impedance Control (IC)}
The FSM-IC is common for prosthesis intrinsic control as studies have shown that humans control the stiffness of leg muscles and therefore joint impedance while walking, and compliant behavior of legs are instrumental for human walking. The impedance controller generates a torque input to the robotic knee based on current knee kinematics and knee joint impedance settings. 

Refer to Fig. \ref{fig:experiment}B, a gait cycle is divided into four phases in the FSM-IC: stance flexion (STF, $m=1$), stance extension (STE, $m=2$), swing flexion (SWF, $m=3$) and swing extension (SWE, $m=4$). The phase transitions are determined by knee motion and gait events (heel strike and toe-off) that are obtained from vertical ground reaction forces of both legs. In each phase of the FSM, three impedance parameters (stiffness $K$, damping $B$, and equilibrium position $\theta_e$) are provided as inputs to the FSM-IC for gait cycle $k$:
\begin{equation}
I^m_k = [K^m_k,B^m_k,(\theta_e)^m_k].
\label{equ:impedance}
\end{equation}
The knee joint torque $T_k^m$ is consequently generated by the following first principle equation,
\begin{equation}
T^m_k=K^m_k(\theta-(\theta_e)^m_k)+B^m_k\omega.
\label{equ:torque}
\end{equation}
The RL controller will adjust these impedance parameters, i.e., 
\begin{equation}
u^m_k = [\Delta K^m_k,\Delta B^m_k,\Delta (\theta_e)^m_k],
\label{actioneq}
\end{equation}
so that the updated impedance parameters are applied to the FSM-IC to generate knee torque according to (\ref{equ:torque}),
\begin{equation}
I^m_{k+1} = I^m_k+u^m_k.
\end{equation}
\begin{figure}
	\centering
	\includegraphics[width=250pt]{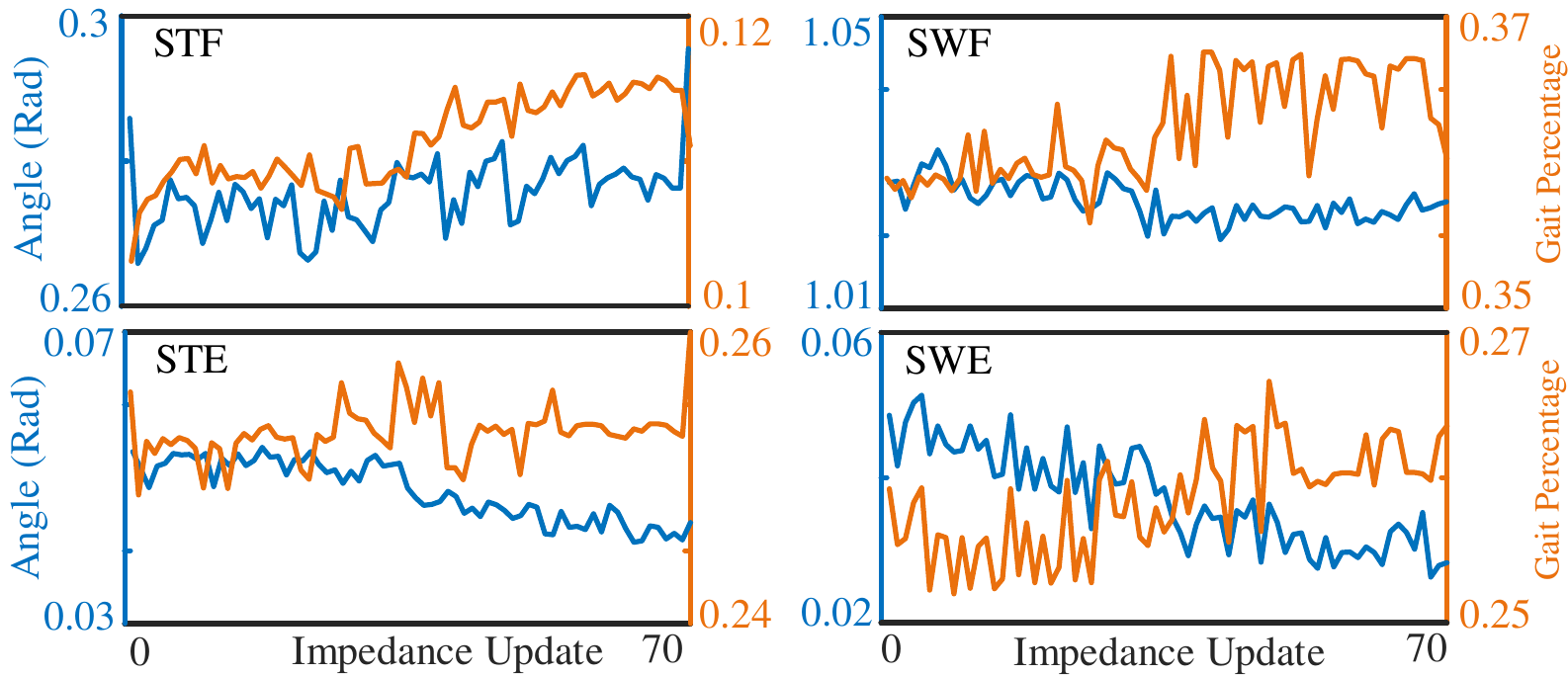}\\
	\caption{An illustration of how the intact knee profile changes as the robotic knee control parameters adapt during a level ground walking simulation session. The peak angles (blue) and the phase durations (orange) are shown in all phases.}
	\label{fig:feature extration}
\end{figure}
\subsection{Tracking Problem Formulation}
Biomechanical studies have shown that the intact knee joint movements or profiles change as amputees adapt to a prosthetic device \cite{huang2015locomotor}. We have observed the same in our pilot study using two human subjects\cite{wu2021reinforcement}. Fig. \ref{fig:feature extration} is an illustration of the same phenomenon using simulations where the intact knee kinematic trajectories were recorded, and changes in profile features are clearly observed. The goal of the RL controlled robotic knee is therefore to track those time varying intact knee profile features for each and every phase during each and every gait cycle. For a gait cycle $k$, the robotic knee motion (Fig. 1B) featured by the peak knee angle $P^m_k$ (degrees) and duration $D^m_k$ (seconds) are measured. Let
\begin{equation}
Z^m_k=({D}^m_ k,{P}^m_k).
\end{equation}
Similarly, we measure the peak knee angle and duration of the intact knee, and let
\begin{equation}
Y_k^m=(\Bar{D}^m_k,\Bar{P}^m_k).
\end{equation}
Consider the human-robot, i.e., the amputee-prosthesis, system as a discrete time nonlinear system with unknown dynamics,
\begin{equation}
Z^m_{k+1}=F({Z}^m_k,{u}^m_k),\ k=0,1,...
\label{equ:DynamicsFunction}
\end{equation}
In (\ref{equ:DynamicsFunction}), the domain of $F(Z_{k}^{m},u_{k}^m)$ is denoted as $\mathcal{D}\triangleq\{(Z^{m},u^{m})|Z\in \mathcal{Z}, u\in \mathcal{U}\}$, where $\mathcal{Z}$ and $\mathcal{U}$ are compact sets with dimensions of $N_Z$ and $N_u$, respectively. In the human-robot system under consideration, $F$ represents the kinematics of the robotic knee, which is affected by both the human wear and also the RL controller. Because of a human in-the-loop, an explicit mathematical model as (\ref{equ:DynamicsFunction}) is intractable or impossible to
obtain.

Without causing any confusion and for the sake of convenience, we drop the superscript $m\ (m=1,2,3,4)$ in the rest of the paper because all four FSM-IC and their respective RL controllers share the same structure, although the RL controllers for each phase have different parameterizations or in other words, the control policies are different for each phase even though they have the same structure.
Then, the tracking error between the intact knee and the prosthetic knee is defined as,
\begin{equation}
e_k=Y_k-Z_k=(\Delta D_k,\Delta P_k).
\end{equation}

\subsection{dHDP for Tracking Control}
Fig. \ref{fig:experiment}A depicts the RL based solution approach to automatically configure the impedance parameters of the robotic knee to track the intact knee joint motion wtihin the FSM-IC framework. Each RL control block corresponds with one of the four FSM phases. As shown in Fig. \ref{fig:experiment}A, we develop a dHDP based RL tracking control with each of the four dHDP blocks providing impedance parameter settings for each of the four gait phases. Each dHDP block has an action network and a critic network, trained for the given FSM phase only. 

In the RL tracking controller, let the state be denoted by $s_{k}$, and the control input/action network output as $u_{k}$ for gait cycle $k$, i.e., 
\begin{equation}
s_k = (\Delta D_k,\Delta P_k), \quad u_k = [\Delta K_k,\Delta B_k,\Delta (\theta_e)_k].
\label{StateEqu}
\end{equation}

We consider the stage cost in a quadratic form 
\begin{equation}
U(s_k,u_k)={s_k}^TR_ss_k + {u_k}^TR_uu_k,
\label{policy1}
\end{equation}
where $R_s\in\mathbb{R}^{2\times2}$ and $R_u\in\mathbb{R}^{3\times3}$ are positive definite matrices.
 
We consider the tracking problem as one to devise an optimal control law via learning from observed data along the human-robot interacting system dynamics. We define the state-action Q-function or the total cost-to-go as, 
\begin{equation}
Q\left(s_{k}, u_{k}\right)=U\left(s_{k}, u_{k}\right)+\sum_{j=1}^{\infty} \gamma^{j} U\left(x_{k+j}, u_{k+j}\right).
\label{Q-function}
\end{equation}

Note that the $Q(s_k,u_k)$ value is a performance measure when action $u_k$ is applied at state $s_k$. Such $Q(s_k,u_k)$ formulation implies that we have considered the optimal adaptive tracking control of the robotic knee as a discrete-time, infinite horizon, discounted problem without knowing an explicit mathematical description of the human-robot interacting dynamics.

For the $Q$-function in (\ref{Q-function}), it satisfies the Bellman equation
\begin{equation}
Q\left(s_{k}, u_{k}\right)=U\left(s_{k}, u_{k}\right)+\gamma Q\left(s_{k+1}, u_{k+1}\right).
\end{equation}
\textbf{Assumption 1.} The state trajectory $Y_{k}$ of the intact knee is bounded, and the initial robotics knee state $Z_{0}$ is bounded.

\noindent\textbf{Remark 1.} Assumption 1 is an intrinsic requirement in biomechanical studies involving human subjects, and it is met with little difficulty as the range of prosthesis knee angles are constrained as shown below.

\subsubsection{Critic Network}
Fig. \ref{fig:experiment}C depicts the structure of the critic network which is realized by an universal approximator with one hidden layer.
Therefore, the approximated value is 
\begin{eqnarray}
\hat{Q}\left(s_{k}, u_{k}\right)=\hat{W}_{c 2,k} \phi(\hat{W}_{c 1,k}z_{k}),
\end{eqnarray}
where $\hat{W}_{c 1,k}$ is the estimated weight matrix between the input layer and the hidden layer, $\hat{W}_{c 2,k}$ is the estimated weight matrix between the hidden layer and the output layer, $\phi$ is the activation function (hyperbolic tangent) in the hidden layer, and the input $z_{k} = [s_{k}, u_{k}]^T$. 

The approximation error of the critic network is
\begin{eqnarray}
e_{c,k} = \gamma \hat{Q}\left(s_{k}, u_{k}\right) - \left[\hat{Q}\left(s_{k-1}, u_{k-1}\right)- U(s_{k-1}, u_{k-1})\right].
\end{eqnarray}

In the following, we use the short-hand notation $U_{k-1}$ for $U(s_{k-1}, u_{k-1})$ and similarly for others. The weights $\hat{W}_{c1}$ and $\hat{W}_{c2}$ are updated as
\begin{eqnarray}
\hat{W}_{c,k+1}=\hat{W}_{c,k}+\Delta W_{c,k},
\end{eqnarray}
according to gradient descend rule as in \cite{si2001online}, $\Delta W_{c1,k}$ and $\Delta W_{c2,k}$ can be written as
\begin{eqnarray}
\begin{aligned}
\Delta \hat{W}_{c1, k}&=-l_{c}\gamma e_{c, k} \hat{W}_{c2, k}\left[\frac{1}{2}\left(1-\phi_{c, k}^{2}\right)\right] z_{k}, \\
\Delta \hat{W}_{c 2,k}&= -l_{c}\gamma e_{c, k} \phi_{c, k},
\label{equ:CNN update}
\end{aligned}
\end{eqnarray}
where $\phi_{c, k}$ is the output of hidden layers in critic network, and $l_{c}>0$ is the learning rate.

\subsubsection{Action network}
The output of the action network is the constrained control input
\begin{eqnarray}
u_{k}=\phi(\hat{W}_{a 2,k} \phi(\hat{W}_{a 1,k}s_{k})),
\end{eqnarray}
where $\hat{W}_{a 1,k}$ is the estimated weight matrix between the input layer and the hidden layer, $\hat{W}_{a 2,k}$ is the estimated weight matrix between the hidden layer and the output layer.

\noindent\textbf{Remark 2.} As hyperbolic tangent is used in the output layer of the actor, the control input $u_{k}$ is constrained.

Based on the design principle of dHDP \cite{si2001online}, the action network is to minimize the total cost-to-go $Q(s_{k}, u_{k})$. We defined the prediction error of the action network as
\begin{eqnarray}
e_{a,k} = \hat{Q}\left(s_{k}, u_{k}\right).
\end{eqnarray}

Similar to (16), the weights $\hat{W}_{a1}$ and $\hat{W}_{a2}$ are updated as
\begin{eqnarray}
\hat{W}_{a,k+1}=\hat{W}_{a,k}+\Delta W_{a,k},
\end{eqnarray}
and $\Delta W_{a1,k}$, $\Delta W_{a2,k}$ can be written as
\begin{eqnarray}
\begin{aligned}
\Delta \hat{W}_{a1,k}&=-l_{a} e_{a,k} \left[\hat{W}_{c2,k} \frac{1}{2}\left(1-\phi_{c, k}^{2}\right) \hat{W}_{cu,k}\right]\frac{1}{2} \left(1-u_{k}^{2}\right) \\
 &\times  \hat{W}_{a2,k}
 \frac{1}{2}\left(1-\phi_{a,k}^{2}\right) s_{k},\\
\Delta \hat{W}_{a2,k} &=-l_{a}e_{a,k} \left[\hat{W}_{c2,k} \frac{1}{2}\left(1-\phi_{c,k}^{2}\right) \hat{W}_{cu,k}\right] \frac{1}{2}\left(1-u_{k}^{2}\right) \phi_{a,k},
\end{aligned}
\label{equ:ANN update}
\end{eqnarray}
where $\hat{W}_{cu,k}$ is the weight vector associated with the input $u_{k}$ (Fig. \ref{fig:experiment}C) from action network, i.e., the part of $\hat{W}_{c1}$ which connects with $u_{k}$, $\phi_{a,k}$ is the output of hidden layers in actor network, and $l_{a}$ is the learning rate.

\section{Lyapunov Stability Analysis}
In this section, we provide a qualitative analysis for the weight convergence of the actor-critic networks, the Bellman (sub)optimality of the control policy, and practical stability of the human-prosthesis system.
\subsection{Preliminaries}
Let $W_{a}^{*}$, $W_{c}^{*}$ denote the optimal weights, that is,
\begin{eqnarray}
\begin{aligned}
W_{a}^{*}=&\arg \min _{\hat{W}_{a}}\Vert\hat{Q}\left(s_{k}, u_{k}\right)\Vert, \\
W_{c}^{*}=&\arg \min _{\hat{W}_{c}}\Vert\gamma \hat{Q}\left(s_{k}, u_{k}\right) + U_{k-1} - \hat{Q}\left(s_{k-1}, u_{k-1}\right)\Vert,
\end{aligned}
\end{eqnarray}
and the optimal $Q$-value and optimal control policy are defined as
\begin{eqnarray}
Q^{*} = W^{*}_{c2}\phi_{c,k} + \epsilon_{c,k}, \quad u^{*} = \phi(W^{*}_{a2}\phi_{a,k}) + \epsilon_{a,k},
\end{eqnarray}
where $\epsilon_{c,k}$ and $\epsilon_{a,k}$ are the reconstruction errors of neural networks.

\noindent\textbf{Assumption 2.} The optimal weights for the actor-critic networks exist and they are bounded by two positive constants $W_{am}$ and $W_{cm}$, respectively,
\begin{eqnarray}
\left\|W_{a}^{*}\right\| \leq W_{am}, \quad \left\|W_{c}^{*}\right\| \leq W_{cm}.	
\end{eqnarray}
Accordingly, the weight estimation errors of the actor-critic networks are described respectively as
\begin{eqnarray}
\tilde{W}_{a,k}:=\hat{W}_{a,k} - W^{*}_{a},\quad  \tilde{W}_{c,k}:=\hat{W}_{c,k} - W^{*}_{c}.
\end{eqnarray} 

\noindent\textbf{Lemma 1.} Under Assumption 2, consider the weight vectors of critic network. Let
\begin{eqnarray}
L_{1,k}=\frac{1}{l_c} tr \left((\tilde{W}_{c2,k})^T \tilde{W}_{c2,k}\right), \quad L_{2,k}=\frac{1}{l_{c} \alpha_{1}} tr\left[\tilde{W}_{c1,k}^{T} \tilde{W}_{c1,k}\right].
\end{eqnarray}
Then the first difference of $L_{1,k}$ is given by
\begin{eqnarray}
\begin{aligned}
\Delta L_{1,k}&= -\gamma^{2}\left\|\zeta_{c,k}\right\|^{2}-\left(1-\gamma^{2} l_{c}\|\phi_{c,k}\right\|^{2}) \\
& \times\|\gamma \hat{W}_{c2,k} \phi_{c,k} + U_{k-1}-\hat{W}_{c2,k-1} \phi_{c,k-1}\|^{2} \\
&  +\|\gamma W_{c2}^{*} \phi_{c,k} + U_{k-1}-\hat{W}_{c2,k-1} \phi_{c,k-1}\|^{2},
\end{aligned}
\end{eqnarray}
where $\zeta_{c,k} =\tilde{W}_{c2,k} \phi_{c,k}$ is an approximation error of the critic output. And the first difference of $L_{2,k}$ is given by
\begin{eqnarray}
\begin{aligned}
\Delta L_{2,k} \leq& \frac{1}{\alpha_{1}}\left(\gamma^{2} l_{c} \| \gamma \hat{W}_{c2,k} \phi_{c,k} + U_{k-1}\right. \\
&-\hat{W}_{c2,k-1} \phi_{c,k-1}\|^{2}\| A_{k}\|^{2}\| x_{c,k} \|^{2} \\
&+\gamma \|\tilde{W}_{c1,k} x_{c,k} A^{T}_{k}\|^{2}+\gamma \| \gamma \hat{W}_{c2,k} \phi_{c,k} \\
&\left. + U_{k-1}-\hat{W}_{c2,k-1} \phi_{c,k-1} \|^{2}\right),
\end{aligned}
\end{eqnarray}
where $\alpha_{1}>0$ is a weighting factor and $A_{k}$ is a vector, with $A_{k}=$ $\frac{1}{2}(1-\phi_{c,k}^{2} ) \hat{W}_{c2,k}$. The proof of Lemma 1 can be found in \cite{yao2020toward}.

\noindent\textbf{Lemma 2.} Under Assumption 2, consider the weight vector of the action network, which realizes the constrained inputs $u_{k}$ to the human-prosthesis systems (7). Let
\begin{eqnarray}
L_{3,k}= \frac{1}{l_{a} \alpha_{2}} tr\left[\tilde{W}_{a2,k}^{T} \tilde{W}_{a2,k}\right], \quad L_{4,k} = \frac{1}{l_{a} \alpha_{3}} tr\left[\tilde{W}_{a1,k}^{T} \tilde{W}_{a1,k} \right].
\end{eqnarray}
Then the first difference of $L_{3,k}$ is bounded by
\begin{eqnarray}
\begin{aligned} 
\Delta L_{3}(k) \leq & \frac{1}{\alpha_{2}}\left(-(1-l_{a}\|\phi_{a2,k}\|^{2}\|\hat{W}_{c2,k} C_{k}\|^{2}) \|\hat{W}_{c2,k} \phi_{c,k} \|^{2}\right.\\ 
&\left.  + 4 \|\zeta_{c,k} \|^{2}+ 4\|W_{c2}^{*} \phi_{c,k} \|^{2} + \|\hat{W}_{c2,k} C_{k} \zeta_{a,k} \|^{2}\right),
\end{aligned}
\end{eqnarray}
where $\zeta_{a,k}=\tilde{W}_{a2,k} \phi_{a,k}$ is an approximation error of the actor network output, $C_{k}=\frac{1}{2}(1-\phi_{c,k}^{2} ) W_{cu,k} \times \frac{1}{2} (1-u_{k}^{2} )$; $\alpha_{2}>0$ is a weighting factor. And the first difference of $L_{4,k}$ is bounded by
\begin{eqnarray}
\begin{aligned} 	
\Delta L_{4}(k) \leq& \frac{1}{\alpha_{3}}\left(l_{a}\|\hat{W}_{c2,k} \phi_{c,k} \|^{2} \|\hat{W}_{c2,k} C_{k} D^{T}_{k}\|^{2} \|s_{k}\|^{2} \right.\\
&\left. + \| \hat{W}_{c2,k} \phi_{c,k} \|^{2} + \|\tilde{W}_{a1,k} s_{k} \|^{2} \|\hat{W}_{c2,k} C_{k} D^{T}_{k} \|^{2}\right),
\end{aligned}
\end{eqnarray}
where $D_{k}=\frac{1}{2}(1-\phi_{a,k}^{2}) \hat{W}_{a2,k}$, and $\alpha_{3}>0$ is a weighting factor.

\noindent\textbf{Proof of Lemma 2.} The first difference of $L_{2,k}$ can be written as
\begin{eqnarray}
\Delta L_{3,k} = \frac{1}{l_{a} \alpha_{2}}tr\left[\tilde{W}_{a2,k+1}^{T} \tilde{W}_{a2,k+1} - \tilde{W}_{a2,k}^{T} \tilde{W}_{a2,k}\right].
\end{eqnarray}
According to (19) and (20), $\tilde{W}_{a2,k+1}$ can be rewritten as
\begin{eqnarray}
\begin{aligned}
\tilde{W}_{a2,k+1}=&\hat{W}_{a2,k+1}-W_{a2}^{*} \\
=& \hat{W}_{a2,k}-l_{a} \phi_{a,k} \hat{W}_{c2,k} C_{k} [\hat{W}_{c2,k} \phi_{c,k}]^{T} - W_{a2}^{*} \\
=& \tilde{W}_{a2,k}-l_{a} \phi_{a,k} \hat{W}_{c2,k} C_{k} [\hat{W}_{c2,k} \phi_{c,k}]^{T}.
\end{aligned}
\end{eqnarray}
Based on the above expression, we can obtain 
\begin{eqnarray}
\begin{aligned}
&tr \left[\tilde{W}_{a2,k+1}^{T} \tilde{W}_{a2,k+1}\right] \\
=&\tilde{W}_{a2,k}^{T} \tilde{W}_{a2,k}+l_{a}^{2}\|\phi_{a,k}\|^{2}\|\hat{W}_{c2,k} C_{k}\|^{2} \|\hat{W}_{c2,k} \phi_{c,k}\|^{2} \\
& -2 l_{a} \hat{W}_{c2,k} C_{k} [\hat{W}_{c2,k} \phi_{c,k}]^{T} \zeta_{a}(k).
\end{aligned}
\end{eqnarray}

Substituting (33) into (31), we have
\begin{eqnarray}
\begin{aligned} 
\Delta L_{3,k}=& \frac{1}{\alpha_{2}}(l_{a}\left\|\phi_{a,k}\right\|^{2}\left\|\hat{W}_{c2,k} C_{k}\right\|^{2} \left\|\hat{W}_{c2,k} \phi_{c,k}\right\|^{2}  \\
&+\| \hat{W}_{c2,k} \phi_{c,k}-\hat{W}_{c2,k} C_{k} \zeta_{a,k} \left\|^{2}-\right\| \hat{W}_{c2,k} C_{k} \zeta_{a,k} \|^{2}\\
&-\left\|\hat{W}_{c2,k} \phi_{c,k}\right\|^{2}).
\end{aligned}
\end{eqnarray}
Note that
\begin{eqnarray}
\begin{aligned}
&\left\|\hat{W}_{c2,k} \phi_{c,k} -\hat{W}_{c2,k} C_{k} \zeta_{a,k} \right\|^{2}-\left\|\hat{W}_{c2,k} C_{k} \zeta_{a,k} \right\|^{2} \\
\leq& 2\left\|\hat{W}_{c2,k} \phi_{c,k} \right\|^{2}+\left\|\hat{W}_{c2,k}  C_{k} \zeta_{a,k} \right\|^{2} \\
\leq& 2\left\|\left(\tilde{W}_{c2,k} + W_{c2}^{*}\right) \phi_{c,k} \right\|^{2}+\left\|\hat{W}_{c2,k} C_{k} \zeta_{a,k} \right\|^{2} \\
\leq& 2\left(\left\|\tilde{W}_{c2,k} \phi_{c,k} \right\|+\left\|W_{c2}^{*} \phi_{c,k} \right\|\right)^{2}+\left\|\hat{W}_{c2,k} C_{k} \zeta_{a,k} \right\|^{2} \\
\leq& 4\left\|\zeta_{c,k} \right\|^{2}+4\left\|W_{c2}^{*} \phi_{c,k}\right\|^{2}+\left\|\hat{W}_{c2,k} C_{k} \zeta_{a,k}\right\|^{2}.
\end{aligned}
\end{eqnarray}
Then we obtain (29) by substituting (35) into (34).

The first difference of $L_{4,k}$ can be written as
\begin{eqnarray}
\Delta L_{4,k} = \frac{1}{l_{a} \alpha_{3}}tr\left[\tilde{W}_{a1,k+1}^{T} \tilde{W}_{a1,k+1} - \tilde{W}_{a1,k}^{T} \tilde{W}_{a1,k}\right].
\end{eqnarray}
According to (19) and (20), $\tilde{W}_{a1,k+1}$ can be rewritten as
\begin{eqnarray}
\begin{aligned}
\tilde{W}_{a1,k+1}=&\hat{W}_{a1,k+1}-W_{a1}^{*}\\
=&\tilde{W}_{a1,k+1} - l_{a} \hat{W}_{c2,k} \phi_{c,k} D_{k} C^{T}_{k} \hat{W}_{c2,k}^{T} s_{k}^{T}.
\end{aligned}
\end{eqnarray}

Let us consider
\begin{eqnarray}
\begin{aligned}
&tr\left[\tilde{W}_{a1,k}^{T} \tilde{W}_{a1,k} \right] \\
=&\tilde{W}_{a1,k}^{T} \tilde{W}_{a1,k} + l_{a}^{2}\left\|\hat{W}_{c2,k} \phi_{c,k} \right\|^{2} \left\|\hat{W}_{c2,k}  C_{k} D^{T}_{k}\right\|^{2}\|s_{k}\|^{2}\\
&-2 l_{a} \hat{W}_{c2,k}  C_{k} D^{T}_{k} \phi_{c,k}^{T} \hat{W}_{c2,k}^{T} \tilde{W}_{a1,k} s_{k}.
\end{aligned}
\end{eqnarray}
Then, by using the cyclic property of matrix trace, the last term in (38) is bounded by 
\begin{eqnarray}
\begin{aligned}
&-2 l_{a} \hat{W}_{c2,k} C_{k} D^{T}_{k} \phi_{c,k}^{T} \left(\hat{W}_{c2,k} \right)^{T} \tilde{W}_{a1,k} s_{k} \\
\leq& l_{a}(\left\|\hat{W}_{c2,k} \phi_{c,k} \right\|^{2} +\left\|\hat{W}_{c2,k} C_{k} D^{T}_{k}\right\|^{2}\left\|\tilde{W}_{a1,k} s_{k}\right\|^{2}). 
\end{aligned}
\end{eqnarray}

Then we obtain (30) by substituting (38), (39) into (36).  $\hfill\blacksquare$ 

\subsection{Weight convergence, (sub)optimality and practical stability}

\noindent\textbf{Definition 1.} (Uniformly ultimately boundedness of a discrete time dynamical system \cite{michel2008stability}) A dynamical system with states $x_{k}$ is said to be uniformly ultimately bounded with ultimate bound $b>0,$ if for any $a>0$ and $t>0,$ there exists a positive number $N=N(a, b)$ independent of $t$, such that $\|\tilde{x}_{k}\| \leq b$ for all $k \geq N+t$ whenever $\left\|\tilde{{x}_{t}}\right\| \leq a$.

In the following, we consider the dynamics of $\hat{W}_{a,k}$ and $\hat{W}_{c,k}$.

\noindent\textbf{Theorem 1.}~(Weight convergence) Under Assumptions 1 and 2, the initial stage cost is bounded. Let the weights of the actor and critic neural networks be updated according to (16) and (20). Then $\tilde{W}_{c}$ and $\tilde{W}_{a}$ are uniformly ultimaltely bounded provided that the following conditions are met:
\begin{eqnarray}
\begin{aligned} 
&l_{c}<\min _{k} \frac{\alpha_{1}-\gamma}{\gamma^{2} \alpha_{1}\left(\left\|\phi_{c,k}\right\|^{2}+\frac{1}{\alpha_{1}}\|A_{k}\|^{2}\|z_{k}\|^{2}\right)}  \\ 
&l_{a}<\min _{k}\left(\alpha_{3}-\alpha_{2}\right)\left(\alpha_{3}\| \hat{W}_{c2,k}^{T} C_{k}\|^{2}\|\phi_{a,k} \|^{2}\right.\\
&\quad \quad \left.+\alpha_{2}\|\hat{W}_{c2,k} C_{k} D^{T}_{k}\|^{2}\|s_{k} \|^{2}\right)^{-1}.
\end{aligned}
\end{eqnarray}

\noindent\textbf{Remark 3.} As $\gamma$, $\alpha_{1}$, $\alpha_{2}$, $\alpha_{3}$ can be found in (11), (25), and (28), for $l_{c}$, $l_{a}$ to be positive, it is necessary that $\alpha_{1}>\gamma>0$ and $\alpha_{3}>\alpha_{2}>0$. Those conditions can be easily satisfied. 

\noindent\textbf{Proof of Theorem 1.} We introduce a candidate Lyapunov function:
\begin{eqnarray}
L_{k} = L_{1,k} + L_{2,k} + L_{3,k} + L_{4,k},
\end{eqnarray}
where $L_{1,k}$, $L_{2,k}$, $L_{3,k}$ and $L_{4,k}$ are defined in (25) and (28). The first difference of $L_{k}$ can be rewritten as
\begin{eqnarray}
\begin{aligned}
\Delta L_{k} \leq &-(\gamma^{2}-\frac{4}{\alpha_{2}})\left\|\zeta_{c,k}\right\|^{2}-(1-\gamma^{2} l_{c}\left\|\phi_{c,k}\right\|^{2}\\
-&\frac{\gamma^{2} l_{c}}{\alpha_{1}}\|A_{k}\|^{2}\|z_{k}\|^{2}-\frac{\gamma}{\alpha_{1}}) \| \gamma \hat{W}_{c2,k} \phi_{c,k} + U_{k-1} \\
-&\hat{W}_{c2,k-1} \phi_{c,k-1} \left\|^{2}-\right\| \hat{W}_{c2,k} \phi_{c,k} \|^{2}(\frac{1}{\alpha_{2}}-\frac{l_{a}}{\alpha_{2}}\left\|\hat{W}_{c2,k} C_{k}\right\|^{2}\\
 \times& \left\|\phi_{a,k}\right\|^{2} -\frac{l_{a}}{\alpha_{3}}\left\|\hat{W}_{c2,k} C_{k} D^{T}_{k}\right\|^{2}\|s_{k}\|^{2}-\frac{1}{\alpha_{3}}) \\
+&\frac{4}{\alpha_{2}}\left\|W_{c2}^{*} \phi_{c,k} \right\|^{2}+\frac{1}{\alpha_{2}}\left\|\hat{W}_{c2,k} C_{k}\right\|^{2}\left\|\zeta_{a,k}\right\|^{2} \\
+&\left\|\gamma W_{c2}^{*} \phi_{c,k} + U_{k-1} - \hat{W}_{c2,k-1} \phi_{c,k-1} \right\|^{2} \\
+&\frac{\gamma}{\alpha_{1}}\left\|\tilde{W}_{c1,k} z_{k}\right\|^{2}\|A_{k}\|^{2} +\frac{1}{\alpha_{3}}\left\|\hat{W}_{c2,k} C_{k} D^{T}_{k}\right\|^{2}\left\|\tilde{W}_{a1,k} s_{k}\right\|^{2}.
\end{aligned}
\end{eqnarray}

To guarantee that the second and the third term in the last expression are negative, we need to choose learning rates in the following manner:
\begin{eqnarray}
1-\gamma^{2} l_{c}\left\|\phi_{c,k} \right\|^{2}-\frac{\gamma^{2} l_{c}}{\alpha_{1}}\|A_{k}\|^{2}\|z_{k}\|^{2}-\frac{\gamma}{\alpha_{1}}>0.
\end{eqnarray}
Therefore,
\begin{eqnarray}
l_{c}<\min _{k} \frac{\alpha_{1}-\gamma}{\gamma^{2} \alpha_{1}(\|\phi_{c,k} \|^{2}+\frac{1}{\alpha_{1}}\|A_{k}\|^{2}\|z_{k}\|^{2})}.
\end{eqnarray}

Similarly, for the action network we obtain
\begin{eqnarray}
\begin{aligned}
&\frac{1}{\alpha_{2}}-\frac{1}{\alpha_{2}} l_{a}\|(\hat{W}_{c2,k} )^{T} C_{k}\|^{2}\|\phi_{a,k}\|^{2} \\
&-\frac{l_{a}}{\alpha_{3}}\|D_{k} C^{T}_{k} \hat{W}_{c2,k} \|^{2}\|s_{k}\|^{2}-\frac{1}{\alpha_{3}}>0,
\end{aligned}
\end{eqnarray}
and then
\begin{eqnarray}
\begin{aligned}
&l_{a}<\min _{k}(\alpha_{3}-\alpha_{2})(\alpha_{3}\|\hat{W}_{c2,k}^{T} C_{k}\|^{2}\|\phi_{a,k}\|^{2}\\
&\quad \quad +\alpha_{2}\|\hat{W}_{c2,k} C_{k} D^{T}_{k}\|^{2}\|s_{k}\|^{2})^{-1}.
\end{aligned}
\end{eqnarray}

As is known, the following inequality holds:
\begin{eqnarray}
\begin{aligned}
&\left\|\gamma W_{c2}^{*} \phi_{c,k} + U_{k-1} - \hat{W}_{c2,k-1} \phi_{c,k-1} \right\|^{2} \\
\leq &4 \gamma^{2}\left\|W_{c2}^{*} \phi_{c,k} \right\|^{2}+4 U_{k-1}^{2} +2\left\|\hat{W}_{c2,k-1} \phi_{c,k-1}\right\|^{2}. 
\end{aligned}
\end{eqnarray}

Applying the Cauchy–Schwarz inequality, we have
\begin{eqnarray}
\begin{aligned}
&\frac{4}{\alpha_{2}} \| W_{c2}^{*} \phi_{c,k} \|^{2} + \frac{1}{\alpha_{2}} \| \hat{W}_{c2,k} C_{k} \|^{2} \| \zeta_{a,k} \|^{2} \\
&+\|\gamma W_{c2}^{*} \phi_{c,k} + U_{k-1} -\hat{W}_{c2,k-1}  \phi_{c,k-1} \|^{2} \\
&+ \frac{\gamma}{\alpha_{1}} \|\tilde{W}_{c1,k} x_{c,k} \|^{2}\|A_{k}\|^{2} +\frac{1}{\alpha_{3}} \|\hat{W}_{c2,k}  C_{k} D^{T}_{k} \|^{2} \|\tilde{W}_{a1,k} s_{k}  \|^{2} \\
\leq &(\frac{4}{\alpha_{2}}+4 \gamma^{2}+2)(W_{2cm} \phi_{cm} )^{2}+4 U_{m}^{2} +\frac{\gamma}{\alpha_{1}}(W_{1cm} z_{m} A_{m})^{2}\\
&+\frac{1}{\alpha_{2}} (W_{2cm} C_{m} W_{2am} \phi_{am})^{2} +\frac{1}{\alpha_{3}}(W_{2cm} C_{m}D_{m} W_{1am} s_{m} )^{2} \\
=&M,
\end{aligned}
\end{eqnarray}
where $\phi_{am}$, $\phi_{cm}$, $U_{m}$, $W_{1cm}$, $W_{2cm}$, $A_{m}$, $C_{m}$, $D_{m}$, $s_{m}$, and  $z_{m}$ are the upper bounds of $\phi_{a,k}$, $\phi_{c,k}$, $U_{k-1}$, $W_{c1,k}$, $W_{c2,k}$, $A_{k}$, $C_{k}$, $D_{k}$, $s_{k}$, and $z_{k}$, respectively.

Substituting (47) and (48) into (42), we obtain $\Delta L_{k}<0$ under the condition that   $\alpha_{2}>\frac{4}{\gamma^{2}}$, constrained $l_{a}$, $l_{c}$ in (40), and
\begin{eqnarray}
\Vert \zeta_{c,k}\Vert> \sqrt{\frac{M}{\gamma^{2}-\frac{4}{\alpha_{2}}}}.
\end{eqnarray}

Given Assumption 1, and that the initial tracking error is bounded, then the stage cost (10) is bounded. From Definition 1, $\Delta L_{k}<0$ means that the estimation errors $\tilde{W}_{a,k}$ and $\tilde{W}_{c,k}$ are bounded from step $k$ to the next step $k+1$, respectively. Established biomechanical principles have provided sufficient conditions on the range of FSM-IC control parameters as safety constraints on the knee joint angles and angular velocities (Remark 1), so that the bounded control law $u_{k}$ results in bounded $Z_{k+1}$, and then we obtain the bounded tracking error $s_{k+1}$. From Remark 2, we know that the control law $u_{k}$ is always constrained at all time steps $k$. Then the resulted stage cost $U_{k+1}$ is bounded. By mathematical induction, we have the estimation errors $\tilde{W}_{a,k}$ and $\tilde{W}_{c,k}$ uniformly ultimately bounded. $\hfill\blacksquare$

\noindent\textbf{Theorem 2.}~((Sub)optimality result) Under the conditions of Theorem 1, the Bellman optimality is achieved within finite approximation error. Meanwhile, the error between the obtained control $u_{k}$ and optimal control $u^{*}$ is uniformly ultimately bounded.

\noindent\textbf{Proof of Theorem 2.} From Assumption 2 and Theorem 1, we obtain that $\hat{Q}(s_{k}, u_{k})$ and $Q^{*}$ are bounded. Furthermore, $u_{k}$ and $u^{*}$ are bounded from Remark 2 and Assumption 2. 

From the approximate $Q$-value (13) and the optimal $Q$-value in (22), we have
\begin{eqnarray}
\begin{aligned}
&\|\hat{Q}(s_{k}, u_{k})-Q^{*} \|\\
=&\| \hat{W}_{c2,k} \phi_{c,k} -W_{c2}^{*} \phi_{c,k}-\epsilon_{c,k} \| \leqslant\|\tilde{W}_{c2,k} \| \phi_{cm}+\epsilon_{cm}.
\end{aligned}
\end{eqnarray}
where $\epsilon_{cm}$ is the upper bound of $\epsilon_{c,k}$.

For $u_{k}$ and $u^{*}$ from (17) and (22), we have
\begin{eqnarray}
\| u_{k}-u^{*} \| = \| \phi(\hat{W}_{a2,k} \phi_{a,k})- \phi(W_{a2}^{*} \phi_{a,k}) - \epsilon_{a,k} \|.
\end{eqnarray}
Similar to (50)
\begin{eqnarray}
\| \hat{W}_{a1,k} \phi_{a2,k}- W_{a2}^{*} \phi_{a2,k} \|  \leqslant  \|\tilde{W}_{a2,k} \| \phi_{a2m}.
\end{eqnarray}
Then, by using the Lagrange Mean Value Theorem, (51) can be rewritten as 
\begin{eqnarray}
\| u_{k}-u^{*} \| \leqslant  \frac{1}{2} \|\tilde{W}_{a2,k} \| \phi_{am} + \epsilon_{am},
\end{eqnarray}
where $\epsilon_{am}$ is the upper bound of $\epsilon_{a,k}$. This comes directly as $\|\tilde{W}_{c2}\|$ and $\|\tilde{W}_{a2}\|$ are both uniformly ultimately bounded as the time step $k$ increases as shown in Theorem 1. It demonstrates that the Bellman optimality is achieved within finite approximation errors. $\hfill\blacksquare$

\noindent\textbf{Remark 4.}~(Practical stability) From Assumption 1, Remarks 1-3 and Theorem 1, we obtain that all the signals (such as $u_k$, $Z_{k}$, $Y_k$ and $s_k$) are bounded in the human-robot system (7). As shown in Theorem 2, the approximated $Q$-value in (13) and the resulted policy (17) achieve (sub)optimality as time step $k$ increases. This demonstrates that the stage cost in (10) approaches zero in the sense of uniformly ultimately boundedness. As such, the tracking error (8) approaches zero in the sense of uniformly ultimately boundedness also as time step $k$ increases. Therefore, the considered human-robot system is practically stable.

\section{OpenSim Simulation Studies}

Table \uppercase\expandafter{\romannumeral1} is a summary of how our proposed dHDP tracking control is implemented and applied to configuring the impedance parameters of the robotic knee. Pertinent information for the implementation steps are provided below. 

\begin{table}
\centering
	\caption{IMPLEMENTATION OF SIMULATION STUDIES}
\flushleft
\begin{tabular}{l}
\hline
Initialization  (gait cycle $k$=0); \\
Set simulation conditions such as walking speed, subject\\
physical conditions (body mass, height, etc.), and small \\
knee angles for both knees;\\
Random, initially feasible impedance parameters;\\
Random initial weights in actor-critic network;\\
\textbf{Repeat} (for a complete OpenSim simulated gait cycle) \\
\quad Obtain target feature $Y_k$ (5) and robotic feature $Z_k$ (6); \\
\quad Obtain state dynamics $s_k$ using (9); \\
\quad \textbf{if} $s_k >$ safety bound \textbf{then} \\
\quad \quad Reset to initial impedance;\\
\quad \quad Repeat the implementation procedure from the top\\ 
\quad \quad but keep the weights of actor-critic networks; \\
\quad \textbf{end if} \\
\quad Obtain constrained input $u_{k}$ through actor network; \\
\quad Obtain stage cost $U_{k}$ using (12);\\
\quad Update critic network using (\ref{equ:CNN update}); \\
\quad Update action network using (\ref{equ:ANN update}); \\
\quad Use $u_{k}$ as input to the FSM-IC by updating the\\ 
\quad impedance  parameters using (4) for the next gait cycle; \\
\textbf{Until} tracking criteria met\\
\hline  
\end{tabular}
\end{table}
\subsection{OpenSim Model Setup}

\begin{figure}
	\centering
	\includegraphics[width=180pt]{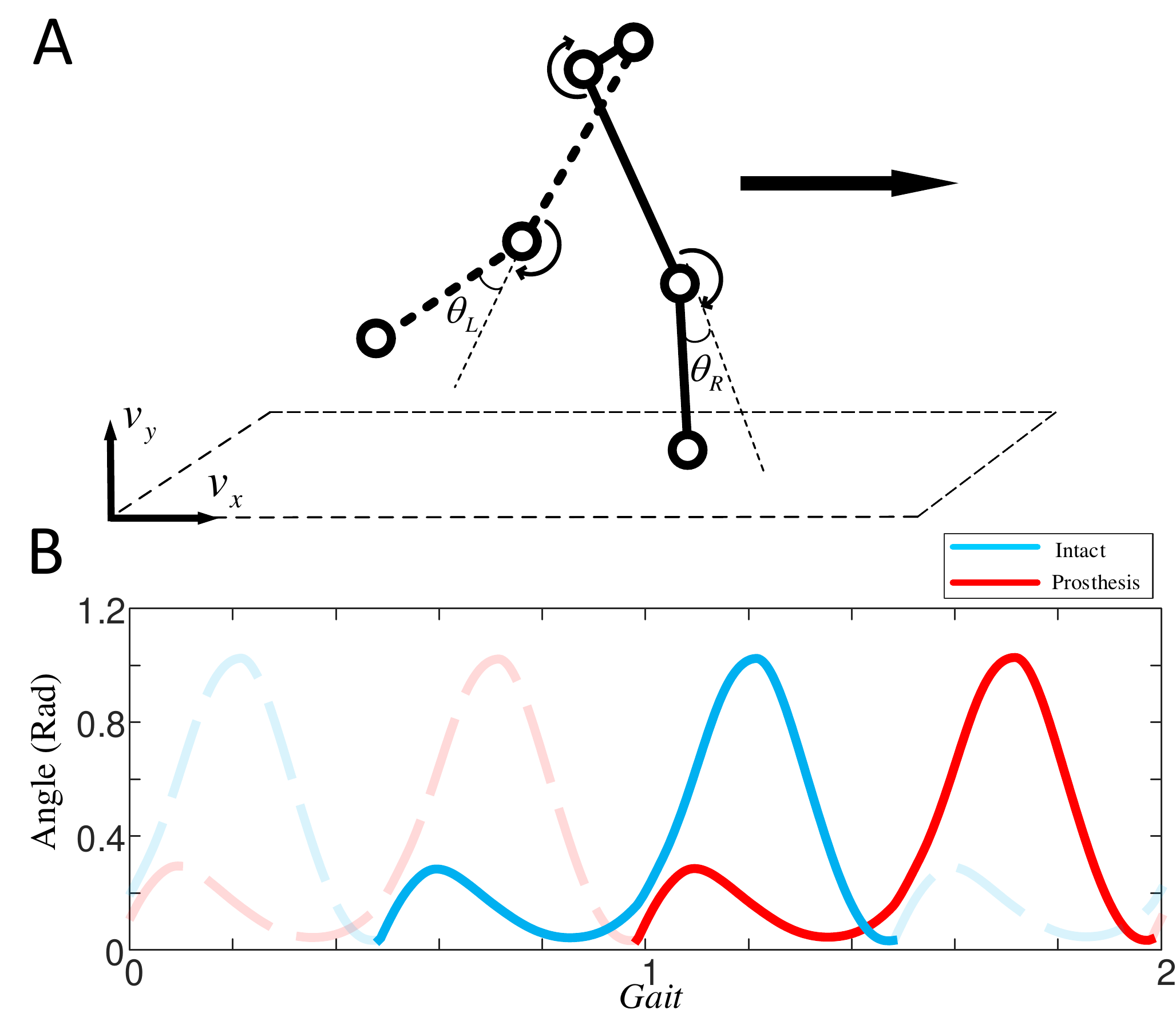}\\
	\caption{OpenSim model. A) Five rigid segment bipedal walking model; B) the next gait is used to measure the tracking error as the prosthetic knee needs to copy the intact knee that is half gait ahead.}
	\label{fig:OpenSim model}
\end{figure}

We investigated this tracking control problem using OpenSim, a well-established simulator in the field of biomechanics \cite{delp2007opensim}. A bipedal walking model, as shown in Fig. \ref{fig:OpenSim model}A, includes a body of five rigid-segments, linked through a one degree of freedom pin joint and the pelvis was linked to the ground by a free joint to allow free movement. The model settings, such as segment length, body mass and inertial parameters, followed the lower limb OpenSim model \cite{jacobs}. In this study, both knees were controlled by FSM-IC controllers. The intact knee was controlled by a fixed set of impedance parameters settings while the prosthetic knee was controlled by the RL controller with its impedance parameters updated for each gait cycle.

Simulating a gait cycle in OpenSim requires specifying initial model settings such as walking speed $v_x$ and $v_y$, both knee angles $\theta_L$ and $\theta_R$ which are set to small numbers near stance position. 

\subsection{ Simulation Procedure}
All simulations were carried out in trials. A trial is a continuous experiment of 500 gait cycles when RL tracking control is applied to tuning the impedance parameters of the robotic knee under different simulation scenarios. We performed two sets of simulations: training trials and testing trials. During training, we performed 30 training trials each from a randomly initialized controller, i.e., a set of randomly generated initially feasible impedance parameter settings, that allows the simulator to simulate balanced walking without falling. During testing, we randomly selected 10 successful controllers (i.e., 10 sets of impedance parameters after training) and applied those control policies (i.e., the actor network weights) as initial controller parameter settings for tracking new, untrained trajectory profiles. Then we tested each of the 10 controllers (policy network weights) to perform 30 new trials, each of which has a new set of randomly selected initial impedance parameters.

A trial is considered successful if the tracking error in (8) reached an error tolerance bound (Table \ref{tab:safety}, bottom row). Specifically, for each of the 4 phases (Fig. \ref{fig:experiment}B), if the tracking error was within the tolerance bound for 8 out of 10 consecutive gait cycles, tracking process in this phase was considered convergent. If all 4 phases had converged within 500 gait cycles, the trial was a success.

To ensure subjects safety (not stumble or fall), a safety bound was introduced based on the realistic conditions of balanced walking. Specifically, as shown in Table II (top row), the safety bound was set at 1.5 standard deviations above the knee kinematic peak values observed in each phase \cite{kadaba1990measurement}. If the tracking errors exceed the safety bound, which means the prosthetic profile may place subjects in unsafe areas, the impedance parameters of the prosthetic knee will be reset to the initial impedance. Note however, the actor and critic network weights are retained for further training until meeting tracking criteria .

In obtaining all results, we set the weighting matrices in the stage cost (10) as: $R_s=diag(1, 1)$ and $R_u=diag(0.1, 0.1, 0.1)$. For the critic network, we used 8 hidden layer neurons with hyperbolic activation function and we used a linear output layer. For the actor network, we used 6 hidden layer neurons with hyperbolic activation function and we also used a hyperbolic activation function for the output layer neuron so that the control inputs are constrained. For both networks, learning rate was 0.1. The actor-critic network was updated every gait cycle until reaching trial success.

\begin{table}[h]
\caption{Safety bound and tolerance bound}
\centering
\setlength{\tabcolsep}{1pt}
 \begin{tabular}{|p{90pt}|p{36pt}|p{36pt}|p{36pt}|p{36pt}|}
 \hline
   & Phase 1 &  Phase 2 & Phase 3& Phase 4 \\ 
 \hline
  Safety Bounds & & & & \\
  {[Angle (rad), Duration ($\%$)]} & [0.184, 12] & [0.131, 12] & [0.157, 12] & [0.105, 12] \\ 
 \hline
 Tolerance Bounds & & & &  \\
 {[Angle (rad), Duration ($\%$)]} &[0.0263, 2]&[0.0263, 2]&[0.0263, 2]&[0.0263, 2] \\
 \hline
  
\end{tabular}
\label{tab:safety}
\end{table}
\subsection{Simulating Realistic Walking}
To evaluate the efficacy of the proposed dHDP tracking control, we performed a systematic simulation study to evaluate human-robot walking performance under RL tracking control. We emulated three walking conditions: ground walking, walking on different terrains and at different paces.

\textbf{\textit{Scenario 1-level ground tracking:}}
The intact knee was operated by a fixed set of impedance parameters at all time while the robotic knee impedance parameters were controlled by RL controller. Note that, fixed impedance parameters still provide realistic control according to (2) as the intact knee was also controlled by FSM-IC. For both the training stage and the testing stage, the RL controller was required to successfully track the intact knee profile within 500 gait cycles. while the OpenSim setting was placed at constant walking pace under level ground condition.

\textbf{\textit{Scenario 2-walking on different terrain:}}
To simulate walking on different terrains, we provided different gait profiles for the intact knee which correspond to different impedance control parameters. The changing profiles of the intact knee were then the new moving targets for the prosthetic knee to track. Five randomly initialized impedance parameter sets that could enable balanced walking of the human-robot system were generated. The impedance parameters of the intact knee were randomly selected from this pool of 5 different gait profiles every 20 gait cycles. During training, the controller was required to successfully track the intact knee for 3 consecutive times with respective knee profiles. During testing, a new pool of 5 sets of impedance parameters was generated and used on the intact knee, the respective profiles of which were tracked by the robotic knee. Again, the controller was required to successfully track the intact knee profile for 3 consecutive times.

\textbf{\textit{Scenario 3-changing pace:}}
We examined RL tracking performance when the pace changes. The simulated pace changes were implemented in a sequence of $[100\%\rightarrow112\%\rightarrow100\%\rightarrow88\%]$ of the initial pace. Changing pace took place once the controllers successfully tracked the intact knee by meeting convergence criteria. The controller must complete the full sequence to complete the training stage. In the testing stage, the pace change was placed in a different order of $[100\%\rightarrow80\%\rightarrow100\%\rightarrow120\%]$ which also signifies a greater variance than the respective conditions for training. Same as in the training stage, each controller must finish the full sequence to be counted as a success.

\begin{figure*}[!ht]
	\centering
	\includegraphics[width=500pt]{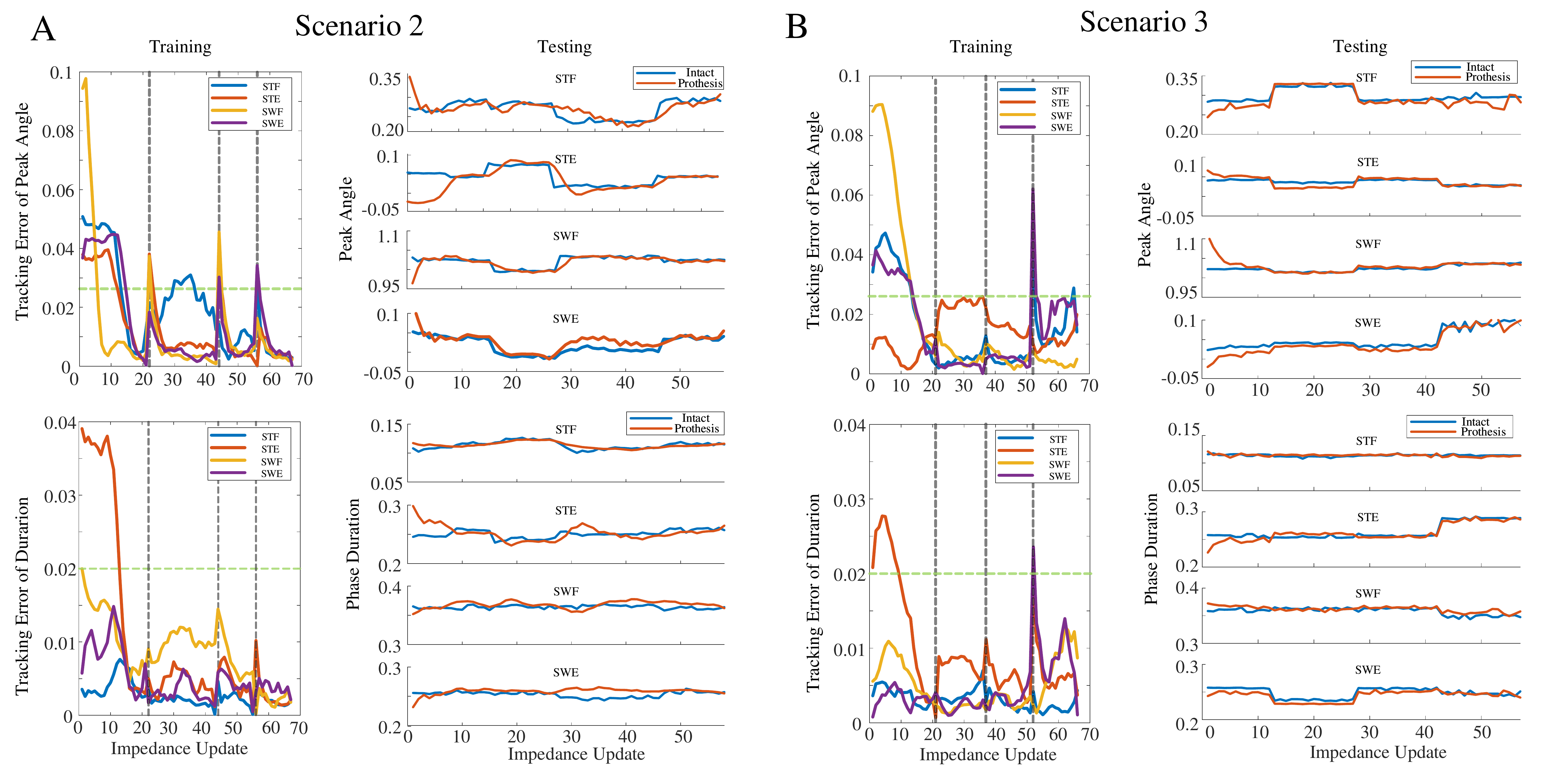}\\
	\caption{Typical trials for scenario 2 (A) and scenario 3 (B). The left columns show tracking errors during training while the horizontal green dashed lines represent tolerance bounds and the grey vertical dashed lines represent task/pace transitions. The right columns show tracking trajectories during testing.}
	\label{fig:scenarios}
\end{figure*}

\section{Results}

\begin{table}[h]
\centering
\caption{ Simulation results of all three scenarios}

\begin{tabular}{|c|c|c|c|} 
\hline
Scenario & Task & Success Rate & Tuning Steps \\ 
\hline
1 & training & 1 & $64.6\pm 84.72$ \\ 
\hline
1 & testing & 1 & $36.30\pm57.22$ \\
\hline
2 & training & 0.97 & $55.76\pm55.18$ \\
\hline
2 & testing & 1 & $19.16\pm15.23$ \\
\hline
3 & training & 0.8 & $52.46\pm32.75$ \\ 
\hline
3 & testing & 0.92 & $42.75\pm37.87$ \\ 
\hline
\end{tabular}
\label{tab:result}
\end{table}

All three scenarios were simulated in both training and testing stages with the same human subject. Table \ref{tab:result} shows the tracking performance for all three scenarios. 

In scenario 1, a $100\%$ success rate was achieved with 64.6 average steps to fully learn to track the intact knee. The RMS tracking error was reduced from 0.0588 to 0.0131 radians of peak angle and from $1.96\%$ to $0.69\%$ of phase duration. In the test stage, because a trained actor network was used in initialization, a improvement in tuning speed was observed from 64.6 to 36.3 average steps. The RMS tracking error has a similar performance.

In scenario 2, a $97\%$ success rate was achieved with 55.8 average steps for successfully track the intact knee each time. The RMS tracking error was reduced from 0.0581 to 0.0124 radians of peak angle and reduced from $1.82\%$ to $0.66\%$ of phase duration. In the test stage, the trained policy of actor network shows the ability to track the target profile rapidly. An average of 19.16 tracking steps with a 100\% success rate shows the ability to track a changing target effectively. Fig. \ref{fig:scenarios}A shows an example of typical training and testing trials. The RMS tracking error has a similar outcome in both training and testing as they share the same success criteria.

Scenario 3 focuses on examining tracking performance reflected in gait duration. $80\%$ success rate was achieved with 52.5 average steps for successfully tracking the intact knee at different paces. The RMS tracking error was reduced from 0.0573 to 0.0103 radians of peak angle and reduced from $1.90\%$ to $0.92\%$ of phase duration. In the test stage, the trained policy of actor network shows a slightly improvement on tracking speed. But it greatly improved the success rate from 0.8 to 0.92. Fig. \ref{fig:scenarios}B shows an example of typical training and testing trial. The RMS tracking error has a similar outcome in both training and testing as they share the same success criteria.

\begin{figure}
	\centering
	\includegraphics[width=250pt]{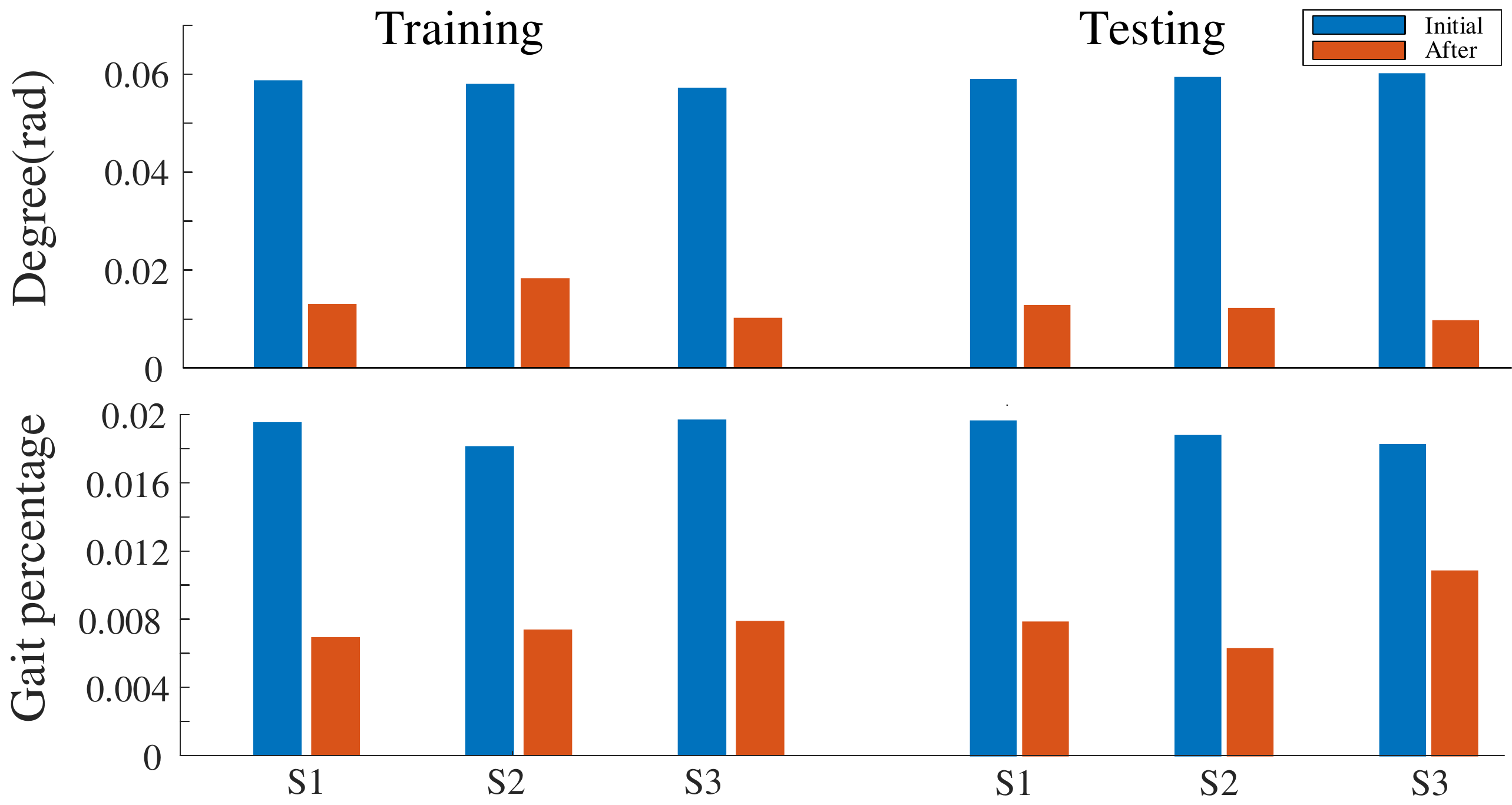}\\
	\caption{Summary of RMS tracking errors of the peak angle (top) and phase duration (bottom) for all three scenarios. The blue bars represent the initial tracking errors while the red bars represent the final tracking errors.}
	\label{fig:FPP result}
\end{figure}

Fig. \ref{fig:FPP result} shows RMS tracking error for all trials in all simulation scenarios. A significant reduction in tracking error was observed under all conditions. The difference between training and testing is minor because they use the same randomization in initial impedance parameters and the same convergence criteria.

\section{Conclusions and Discussions}
We have introduced a new RL based tracking control scheme for automatic tuning of robotic knee impedance parameter settings of a robotic knee to track the intact knee kinematics. For the first time, we successfully demonstrated stable and continuous walking in simulations of a human wearing a robotic prosthesis which was designed to track the intact knee motion. 

Mirroring the intact knee motion by a prosthetic knee is an intuitive idea which has been proposed for decades, but have not been successfully demonstrated. The robotic knee control to mimic the intact knee joint is a tracking problem in classical control. Even though many control theoretic solutions exist, such as backstepping\cite{krstic1995nonlinear,jiangdagger1997tracking,do2004simultaneous}, observer-based control\cite{khalil2002nonlinear} and nonlinear adaptive/robust control\cite{nijmeijer1990nonlinear,isidori2013nonlinear}, they are inadequate for this problem as they require an accurate mathematical description of the system dynamics, which involve co-adapting human and robot in this case, and which are nearly impossible to obtain. Additionally, those control theoretic approaches focus on the stabilization (in Lyapunov sense) of the nonlinear dynamic systems without addressing control performance such as Bellman optimality.

Recently, some results emerged to tackle these issues using data-driven, learning enabled, nonlinear optimal tracking control designs\cite{hou2013model}.  Unfortunately, many of the reported results have focused on theoretical analyses, which are usually based on requiring a reference model for the desired movement trajectory and/or control trajectory. They are thus not practically useful.

In this paper, we have presented a complete tracking control algorithm based on dHDP, and an analytical framework for the tracking controller with constrained inputs. Additionally, we have systematically evaluated the performance of the proposed tracking controller. Our simulation results have shown effectiveness of the tracking controller for different walking tasks that emulate level ground walking, walking on different terrains and at different paces. Based on our previous work, we expect these new results to be verified in human experiments at a future time.

\bibliographystyle{IEEEtran}

\bibliography{ICRA}

\end{document}